\documentclass[11pt,a4paper]{article}
\usepackage[hyperref]{emnlp2017}
\usepackage{times}
\usepackage{latexsym}
\usepackage{hyperref}

\usepackage{amsmath, amssymb}
\usepackage{bm}
\usepackage[pdftex]{graphicx}
\usepackage{url}
\usepackage{multirow}

\usepackage{hasen}

\emnlpfinalcopy 

\title{Neural Machine Translation with Source-Side Latent Graph Parsing}

\author{
	Kazuma Hashimoto and Yoshimasa Tsuruoka\\
	The University of Tokyo, 7-3-1 Hongo, Bunkyo-ku, Tokyo, Japan\\
	{\tt \{hassy,tsuruoka\}@logos.t.u-tokyo.ac.jp}\\
}

\date{}

\begin{document}

\maketitle

\begin{abstract}
This paper presents a novel neural machine translation model which jointly learns translation and source-side latent graph representations of sentences.
Unlike existing pipelined approaches using syntactic parsers, our end-to-end model learns a latent graph parser as part of the encoder of an attention-based neural machine translation model, and thus the parser is optimized according to the translation objective.
In experiments, we first show that our model compares favorably with state-of-the-art sequential and pipelined syntax-based NMT models.
We also show that the performance of our model can be further improved by pre-training it with a small amount of treebank annotations.
Our final ensemble model significantly outperforms the previous best models on the standard English-to-Japanese translation dataset.
\end{abstract}

\section{Introduction}
Neural Machine Translation (NMT) is an active area of research due to its outstanding empirical results~\citep{attention,luong2015,seq2seq}.
Most of the existing NMT models treat each sentence as a sequence of tokens, but recent studies suggest that syntactic information can help improve translation accuracy~\citep{eriguchi2016,eriguchi2017,sennrich_wmt,guide}.
The existing syntax-based NMT models employ a syntactic parser trained by supervised learning in advance, and hence the parser is not adapted to the translation tasks.
An alternative approach for leveraging syntactic structure in a language processing task is to jointly learn syntactic trees of  the sentences along with the target task~\citep{socher2011,deepmind}.

\begin{figure}[t]
	\begin{center}
    	\includegraphics[width=75mm]{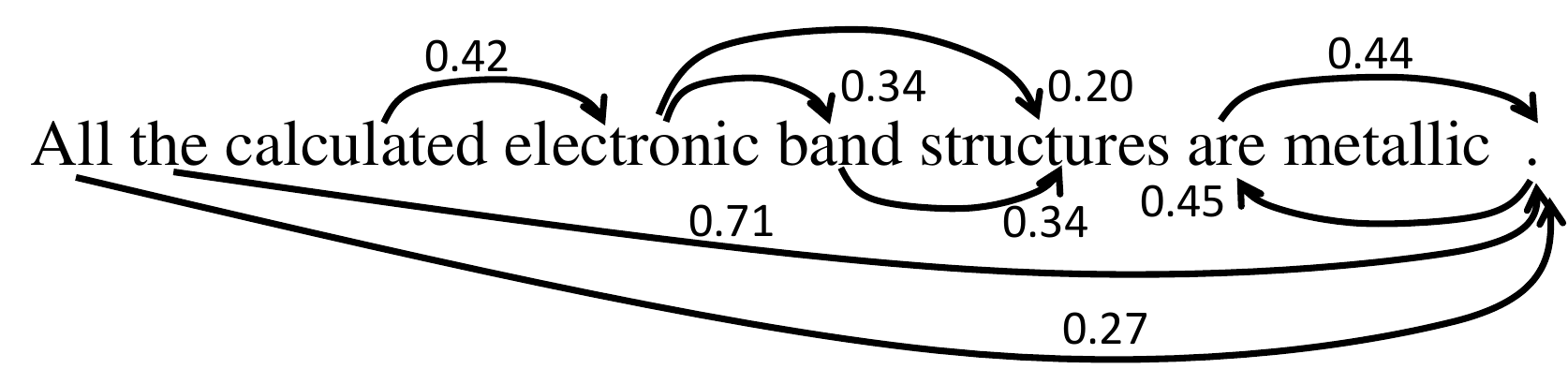}
    \end{center}
\vspace{-0.5cm}
\caption{An example of the learned latent graphs. Edges with a small weight are omitted.}
\label{fig:overview}
\end{figure}

Motivated by the promising results of recent joint learning approaches, we present a novel NMT model that can learn a task-specific latent graph structure for each source-side sentence.
The graph structure is similar to the dependency structure of the sentence, but it can have cycles and is learned specifically for the translation task.
Unlike the aforementioned approach of learning single syntactic trees, our latent graphs are composed of ``soft'' connections, i.e., the edges have real-valued weights (Figure~\ref{fig:overview}).
Our model consists of two parts: one is a task-independent parsing component, which we call a {\it latent graph parser}, and the other is an attention-based NMT model.
The latent parser can be independently pre-trained with human-annotated treebanks and is then adapted to the translation task.

In experiments, we demonstrate that our model can be effectively pre-trained by the treebank annotations, outperforming a state-of-the-art sequential counterpart and a pipelined syntax-based model.
Our final ensemble model outperforms the previous best results by a large margin on the WAT English-to-Japanese dataset.

\section{Latent Graph Parser}
\label{sec:graph}

We model the latent graph parser based on dependency parsing.
In dependency parsing, a sentence is represented as a tree structure where each node corresponds to a word in the sentence and a unique {\it root} node (ROOT) is added.
Given a sentence of length $N$, the parent node $H_{w_i}\in\{w_1,\ldots,w_N, \mathrm{ROOT}\}$ $(H_{w_i}\neq w_i)$ of each word $w_i$ $(1\leq i \leq N)$ is called its {\it head}.
The sentence is thus represented as a set of tuples $(w_i, H_{w_i}, \ell_{w_i})$, where $\ell_{w_i}$ is a dependency label.

In this paper, we remove the constraint of using the tree structure and represent a sentence as a set of tuples $(w_i, p(H_{w_i}|w_i), p(\ell_{w_i}|w_i))$, where $p(H_{w_i}|w_i)$ is the probability distribution of $w_i$'s parent nodes, and $p(\ell_{w_i}|w_i)$ is the probability distribution of the dependency labels.
For example, $p(H_{w_i}=w_j|w_i)$ is the probability that $w_j$ is the parent node of $w_i$.
Here, we assume that a special token $\langle$EOS$\rangle$ is appended to the end of the sentence, and we treat the $\langle$EOS$\rangle$ token as ROOT.
This approach is similar to that of graph-based dependency parsing~\citep{mc2005} in that a sentence is represented with a set of weighted arcs between the words.
To obtain the {\it latent graph} representation of the sentence, we use a dependency parsing model based on multi-task learning proposed by \citet{jmt_emnlp}.

\subsection{Word Representation}
The $i$-th input word $w_i$ is represented with the concatenation of its $d_1$-dimensional word embedding $v_{dp}(w_i)\in\mathbb{R}^{d_1}$ and its character $n$-gram embedding $c(w_i)\in\mathbb{R}^{d_1}$: $x(w_i)=[v_{dp}(w_i); c(w_i)]$.
$c(w_i)$ is computed as the average of the embeddings of the character $n$-grams in $w_i$.

\subsection{POS Tagging Layer}
Our latent graph parser builds upon multi-layer bi-directional Recurrent Neural Networks (RNNs) with Long Short-Term Memory (LSTM) units~\citep{lstm2}.
In the first layer, POS tagging is handled by computing a hidden state $h^{(1)}_i=[\overrightarrow{h}^{(1)}_i; \overleftarrow{h}^{(1)}_i]\in\mathbb{R}^{2d_1}$ for $w_i$, where $\overrightarrow{h}^{(1)}_i=\mathrm{LSTM}(\overrightarrow{h}^{(1)}_{i-1}, x(w_i))\in\mathbb{R}^{d_1}$ and $\overleftarrow{h}^{(1)}_i=\mathrm{LSTM}(\overleftarrow{h}^{(1)}_{i+1}, x(w_i))\in\mathbb{R}^{d_1}$ are hidden states of the forward and backward LSTMs, respectively.
$h^{(1)}_i$ is then fed into a softmax classifier to predict a probability distribution $p^{(1)}_i\in\mathbb{R}^{C^{(1)}}$ for word-level tags, where $C^{(1)}$ is the number of POS classes.
The model parameters of this layer can be learned not only by human-annotated data, but also by backpropagation from higher layers, which are described in the next section.

\subsection{Dependency Parsing Layer}
Dependency parsing is performed in the second layer.
A hidden state $h^{(2)}_i\in\mathbb{R}^{2d_1}$ is computed by $\overrightarrow{h}^{(2)}_i=\mathrm{LSTM}(\overrightarrow{h}^{(2)}_{i-1}, [x(w_i); y(w_i); \overrightarrow{h}^{(1)}_{i}])$ and $\overleftarrow{h}^{(2)}_i=\mathrm{LSTM}(\overleftarrow{h}^{(2)}_{i+1}, [x(w_i); y(w_i); \overleftarrow{h}^{(1)}_{i}])$, where $y(w_i)=W^{(1)}_{\ell}p^{(1)}_i\in\mathbb{R}^{d_2}$ is the POS information output from the first layer, and $W^{(1)}_{\ell}\in\mathbb{R}^{d_2\times C^{(1)}}$ is a weight matrix.

Then, (soft) edges of our latent graph representation are obtained by computing the probabilities
\begin{equation}
p(H_{w_i}=w_j|w_i)=\frac{\exp{(m(i, j))}}{\sum_{k\neq i}{\exp{(m(i, k))}}},
\end{equation}
where $m(i, k)=h^{(2)\mathrm{T}}_{k}W_{dp}h^{(2)}_i$ $(1\leq k \leq N+1, k\neq i)$ is a scoring function with a weight matrix $W_{dp}\in\mathbb{R}^{2d_1\times 2d_1}$.
While the models of \citet{jmt_emnlp}, \citet{head_selec}, and \citet{biaffine2017} learn the model parameters of their parsing models only by human-annotated data, we allow the model parameters to be learned by the translation task.

Next, $[h^{(2)}_i; z(H_{w_i})]$ is fed into a softmax classifier to predict the probability distribution $p(\ell_{w_i}|w_i)$, where $z(H_{w_i})\in\mathbb{R}^{2d_1}$ is the weighted average of the hidden states of the parent nodes: $\sum_{j\neq i}p(H_{w_i}=w_j|w_i)h^{(2)}_j$.
This results in the latent graph representation $(w_i, p(H_{w_i}|w_i), p(\ell_{w_i}|w_i))$ of the input sentence.

\section{NMT with Latent Graph Parser}
The latent graph representation described in Section~\ref{sec:graph} can be used for any sentence-level tasks, and here we apply it to an Attention-based NMT (ANMT) model~\citep{luong2015}.
We modify the encoder and the decoder in the ANMT model to learn the latent graph representation.

\subsection{Encoder with Dependency Composition}
The ANMT model first encodes the information about the input sentence and then generates a sentence in another language.
The encoder represents the word $w_i$ with a word embedding $v_{enc}(w_i)\in\mathbb{R}^{d_3}$.
It should be noted that $v_{enc}(w_i)$ is different from $v_{dp}(w_i)$ because each component is separately modeled.
The encoder then takes the word embedding $v_{enc}(w_i)$ and the hidden state $h^{(2)}_i$ as the input to a uni-directional LSMT:
\begin{equation}
h^{(enc)}_i=\mathrm{LSTM}(h^{(enc)}_{i-1}, [v_{enc}(w_i); h^{(2)}_i]),
\end{equation}
where $h^{(enc)}_i\in\mathbb{R}^{d_3}$ is the hidden state corresponding to $w_i$.
That is, the encoder of our model is a three-layer LSTM network, where the first two layers are bi-directional.

In the sequential LSTMs, relationships between words in distant positions are not {\it explicitly} considered.
In our model, we explicitly incorporate such relationships into the encoder by defining a dependency composition function:
\begin{equation}
\label{eq:depComp}
dep(w_i)=\mathrm{tanh}(W_{dep}[h^{enc}_i; \overline{h}(H_{w_i});p(\ell_{w_i}|w_i)]),
\end{equation}
where $\overline{h}(H_{w_i})=\sum_{j\neq i}p(H_{w_i}=w_j|w_i)h^{(enc)}_j$ is the weighted average of the hidden states of the parent nodes.

\paragraph{Note on character $n$-gram embeddings}
In NMT models, sub-word units are widely used to address rare or unknown word problems~\citep{bpe}.
In our model, the character $n$-gram embeddings are fed through the latent graph parsing component.
To the best of our knowledge, the character $n$-gram embeddings have never been used in NMT models.
\citet{wieting_char2016}, \citet{char2017}, and \citet{jmt_emnlp} have reported that the character $n$-gram embeddings are useful in improving several NLP tasks by better handling unknown words.

\subsection{Decoder with Attention Mechanism}
The decoder of our model is a single-layer LSTM network, and the initial state is set with $h^{(enc)}_{N+1}$ and its corresponding memory cell.
Given the $t$-th hidden state $h^{(dec)}_t\in\mathbb{R}^{d_3}$, the decoder predicts the $t$-th word in the target language using an attention mechanism.
The attention mechanism in \citet{luong2015} computes the weighted average of the hidden states $h^{(enc)}_i$ of the encoder:
\begin{eqnarray}
s(i, t)=&\frac{\exp{(h^{(dec)}_t\cdot h^{(enc)}_i)}}{\sum_{j=1}^{N+1}\exp{(h^{(dec)}_t\cdot h^{(enc)}_j)}},\label{eq:attn_score}\\
a_t=&\sum_{i=1}^{N+1}s(i, t)h^{(enc)}_i,
\end{eqnarray}
where $s(i, t)$ is a scoring function which specifies how much each source-side hidden state contributes to the word prediction.

In addition, like the attention mechanism over constituency tree nodes~\citep{eriguchi2016}, our model uses attention to the dependency composition vectors:
\begin{eqnarray}
s'(i, t)=&\frac{\exp{(h^{(dec)}_t\cdot dep(w_i))}}{\sum_{j=1}^{N}\exp{(h^{(dec)}_t\cdot dep(w_j))}},\\
a'_t=&\sum_{i=1}^{N}s'(i, t)dep(w_i),\label{eq:depAttn}
\end{eqnarray}
To predict the target word, a hidden state $\tilde{h}^{(dec)}_t\in\mathbb{R}^{d_3}$ is then computed as follows:
\begin{equation}
\label{eq:s_tilde}
\tilde{h}^{(dec)}_t=\mathrm{tanh}(\tilde{W}[h^{(dec)}_t; a_t; a'_t]),
\end{equation}
where $\tilde{W}\in\mathbb{R}^{d_3\times 3d_3}$ is a weight matrix.
$\tilde{h}^{(dec)}_t$ is fed into a softmax classifier to predict a target word distribution.
$\tilde{h}^{(dec)}_t$ is also used in the transition of the decoder LSTMs along with a word embedding $v_{dec}(w_t)\in\mathbb{R}^{d_3}$ of the target word $w_t$:
\begin{equation}
h^{(dec)}_{t+1}=\mathrm{LSTM}(h^{(dec)}_t, [v_{dec}(w_t); \tilde{h}^{(dec)}_t]),
\end{equation}
where the use of $\tilde{h}^{(dec)}_t$ is called {\it input feeding} proposed by \citet{luong2015}.

The overall model parameters, including those of the latent graph parser, are jointly learned by minimizing the negative log-likelihood of the prediction probabilities of the target words in the training data.
To speed up the training, we use BlackOut sampling~\citep{blackout}.
By this joint learning using Equation~(\ref{eq:depComp}) and (\ref{eq:depAttn}), the latent graph representations are automatically learned according to the target task.

\paragraph{Implementation Tips}
Inspired by \citet{nce_gpu}, we further speed up BlackOut sampling by sharing noise samples across words in the same sentences.
This technique has proven to be effective in RNN language modeling, and we have found that it is also effective in the NMT model.
We have also found it effective to share the model parameters of the target word embeddings and the softmax weight matrix for word prediction~\citep{tai_softmax,tai_softmax_2}.
Also, we have found that a parameter averaging technique~\cite{hashimoto2013} is helpful in improving translation accuracy.

\paragraph{Translation}
At test time, we use a novel beam search algorithm which combines statistics of sentence lengths~\citep{eriguchi2016} and length normalization~\citep{length}.
During the beam search step, we use the following scoring function for a generated word sequence $y=(y_1, y_2,\ldots, y_{L_y})$ given a source word sequence $x=(x_1, x_2, \ldots, x_{L_x})$:
\begin{equation}
\frac{1}{L_y}\left(\sum_{i=1}^{L_y}\log{p(y_i|x, y_{1:i-1})}+\log{p(L_y|L_x)}\right),
\end{equation}
where $p(L_y|L_x)$ is the probability that sentences of length $L_y$ are generated given source-side  sentences of length $L_x$.
The statistics are taken by using the training data in advance.
In our experiments, we have empirically found that this beam search algorithm helps the NMT models to avoid generating translation sentences that are too short.

\section{Experimental Settings}

\subsection{Data}
We used an English-to-Japanese translation task of the Asian Scientific Paper Excerpt Corpus (ASPEC)~\citep{aspec} used in the Workshop on Asian Translation (WAT), since it has been shown that syntactic information is useful in English-to-Japanese translation~\citep{eriguchi2016,neubig2015}.
We followed the data preprocessing instruction for the English-to-Japanese task in \citet{eriguchi2016}.
The English sentences were tokenized by the tokenizer in the Enju parser~\citep{enju}, and the Japanese sentences were segmented by the KyTea tool\footnote{\url{http://www.phontron.com/kytea/}.}.
Among the first 1,500,000 translation pairs in the training data, we selected 1,346,946 pairs where the maximum sentence length is 50.
In what follows, we call this dataset the {\it large} training dataset.
We further selected the first 20,000 and 100,000 pairs to construct the {\it small} and {\it medium} training datasets, respectively.
The development data include 1,790 pairs, and the test data 1,812 pairs.

For the small and medium datasets, we built the vocabulary with words whose minimum frequency is two, and for the large dataset, we used words whose minimum frequency is three for English and five for Japanese.
As a result, the vocabulary of the target language was 8,593 for the small dataset, 23,532 for the medium dataset, and 65,680 for the large dataset.
A special token $\langle$UNK$\rangle$ was used to replace words which were not included in the vocabularies.
The character $n$-grams ($n=2, 3, 4$) were also constructed from each training dataset with the same frequency settings.

\subsection{Parameter Optimization and Translation}
We turned hyper-parameters of the model using development data.
We set $(d_1, d_2)=(100, 50)$ for the latent graph parser.
The word and character $n$-gram embeddings of the latent graph parser were initialized with the pre-trained embeddings in \citet{jmt_emnlp}.\footnote{The pre-trained embeddings can be found at \url{https://github.com/hassyGo/charNgram2vec}.}
The weight matrices in the latent graph parser were initialized with uniform random values in $[-\frac{\sqrt{6}}{\sqrt{row+col}}, +\frac{\sqrt{6}}{\sqrt{row+col}}]$, where $row$ and $col$ are the number of rows and columns of the matrices, respectively.
All the bias vectors and the weight matrices in the softmax layers were initialized with zeros, and the bias vectors of the forget gates in the LSTMs were initialized by ones~\citep{bias1}.

We set $d_3=128$ for the small training dataset, $d_3=256$ for the medium training dataset, and $d_3=512$ for the large training dataset.
The word embeddings and the weight matrices of the NMT model were initialized with uniform random values in $[-0.1, +0.1]$.
The training was performed by mini-batch stochastic gradient descent with momentum.
For the BlackOut objective~\citep{blackout}, the number of the negative samples was set to 2,000 for the small and medium training datasets, and 2,500 for the large training dataset.
The mini-batch size was set to 128, and the momentum rate was set to 0.75 for the small and medium training datasets and 0.70 for the large training dataset.
A gradient clipping technique was used with a clipping value of 1.0.
The initial learning rate was set to 1.0, and the learning rate was halved when translation accuracy decreased.
We used the BLEU scores obtained by greedy translation as the translation accuracy and checked it at every half epoch of the model training.
We saved the model parameters at every half epoch and used the saved model parameters for the parameter averaging technique.
For regularization, we used L2-norm regularization with a coefficient of $10^{-6}$ and applied dropout~\citep{dropout} to Equation~(\ref{eq:s_tilde}) with a dropout rate of 0.2.

The beam size for the beam search algorithm was 12 for the small and medium training datasets, and 50 for the large training dataset.
We used BLEU~\citep{bleu}, RIBES~\citep{ribes}, and perplexity scores as our evaluation metrics.
Note that lower perplexity scores indicate better accuracy.

\subsection{Pre-Training of Latent Graph Parser}
\label{subsec:pre-training}
The latent graph parser in our model can be optionally pre-trained by using human annotations for dependency parsing.
In this paper we used the widely-used Wall Street Journal (WSJ) training data to jointly train the POS tagging and dependency parsing components.
We used the standard training split (Section 0-18) for POS tagging.
We followed \citet{chen2014dep} to generate the training data (Section 2-21) for dependency parsing.
From each training dataset, we selected the first $K$ sentences to pre-train our model.
The training dataset for POS tagging includes 38,219 sentences, and that for dependency parsing includes 39,832 sentences.

The parser including the POS tagger was first trained for 10 epochs in advance according to the multi-task learning procedure of \citet{jmt_emnlp}, and then the overall NMT model was trained.
When pre-training the POS tagging and dependency parsing components, we did not apply dropout to the model and did not fine-tune the word and character $n$-gram embeddings to avoid strong overfitting.

\subsection{Model Configurations}
\paragraph{LGP-NMT}
is our proposed model that learns the Latent Graph Parsing for NMT.

\paragraph{LGP-NMT+}
is constructed by pre-training the latent parser in LGP-NMT as described in Section~\ref{subsec:pre-training}.

\paragraph{SEQ}
is constructed by removing the dependency composition in Equation~(\ref{eq:depComp}), forming a sequential NMT model with the multi-layer encoder.

\paragraph{DEP} is constructed by using pre-trained dependency relations rather than learning them.
That is, $p(H_{w_i}=w_j|w_i)$ is fixed to 1.0 such that $w_j$ is the head of $w_i$.
The dependency labels are also given by the parser which was trained by using all the training samples for parsing and tagging.

\paragraph{UNI}
is constructed by fixing $p(H_{w_i}=w_j|w_i)$ to $\frac{1}{N}$ for all the words in the same sentence.
That is, the uniform probability distributions are used for equally connecting all the words.


\section{Results on Small and Medium Datasets}

We first show our translation results using the small and medium training datasets.
We report averaged scores with standard deviations across five different runs of the model training.

\subsection{Small Training Dataset}

\begin{table}
  \begin{center}
{\small
    \begin{tabular}{l|ccc}
						& BLEU & RIBES & Perplexity \\ \hline
	LGP-NMT  		& 14.31$\pm$1.49  & 65.96$\pm$1.86 & 41.13$\pm$2.66  \\
	LGP-NMT+	     & 16.81$\pm$0.31 &  69.03$\pm$0.28 & 38.33$\pm$1.18 \\ \hdashline
	SEQ	 			& 15.37$\pm$1.18 & 67.01$\pm$1.55 & 38.12$\pm$2.52 \\
	UNI		 			& 15.13$\pm$1.67 & 66.95$\pm$1.94 & 39.25$\pm$2.98 \\
	DEP	 			& 13.34$\pm$0.67 & 64.95$\pm$0.75 & 43.89$\pm$1.52 \\ \hline
    \end{tabular}
}
    \caption{Evaluation on the development data using the small training dataset (20,000 pairs).}
    \label{tb:comparison_small}
  \end{center}
%
  \begin{center}
{\small
    \begin{tabular}{r|ccc}
	$K$		& BLEU & RIBES & Perplexity \\ \hline
	0  			& 14.31$\pm$1.49  & 65.96$\pm$1.86 & 41.13$\pm$2.66  \\
	5,000	     & 16.99$\pm$1.00 &  69.03$\pm$0.93 & 37.14$\pm$1.96 \\
	10,000		& 16.81$\pm$0.31 & 69.03$\pm$0.28 & 38.33$\pm$1.18 \\
	All 			& 16.09$\pm$0.56 &  68.19$\pm$0.59 & 39.24$\pm$1.88 \\ \hline
    \end{tabular}
}
    \caption{Effects of the size $K$ of the training datasets for POS tagging and dependency parsing.}
    \label{tb:comparison_small_dep}
  \end{center}
\end{table}

Table~\ref{tb:comparison_small} shows the results of using the small training dataset.
LGP-NMT performs worse than SEQ and UNI, which shows that the small training dataset is not enough to learn useful latent graph structures from scratch.
However, LGP-NMT+ ($K$ = 10,000) outperforms SEQ and UNI, and the standard deviations are the smallest.
Therefore, the results suggest that pre-training the parsing and tagging components can improve the translation accuracy of our proposed model.
We can also see that DEP performs the worst.
This is not surprising because previous studies, e.g., \citet{when_tree}, have reported that using syntactic structures do not always outperform competitive sequential models in several NLP tasks.

Now that we have observed the effectiveness of pre-training our model, one question arises naturally:
\begin{itemize}
\item[]how many training samples for parsing and tagging are necessary for improving the translation accuracy?
\end{itemize}
Table~\ref{tb:comparison_small_dep} shows the results of using different numbers of training samples for parsing and tagging.
The results of $K$= 0 and $K$= 10,000 correspond to those of LGP-NMT and LGP-NMT+ in Table~\ref{tb:comparison_small}, respectively.
We can see that using the small amount of the training samples performs better than using all the training samples.\footnote{We did not observe such significant difference when using the larger datasets, and we used all the training samples in the remaining part of this paper.}
One possible reason is that the domains of the translation dataset and the parsing (tagging) dataset are considerably different.
The parsing and tagging datasets come from WSJ, whereas the translation dataset comes from abstract text of scientific papers in a wide range of domains, such as biomedicine and computer science.
These results suggest that our model can be improved by a small amount of parsing and tagging datasets in different domains.
Considering the recent universal dependency project\footnote{\url{http://universaldependencies.org/}.} which covers more than 50 languages, our model has the potential of being applied to a variety of language pairs.

\subsection{Medium Training Dataset}

\begin{table}
  \begin{center}
{\small
    \begin{tabular}{l|ccc}
						& BLEU & RIBES & Perplexity \\ \hline
	LGP-NMT  		& 28.70$\pm$0.27  & 77.51$\pm$0.13 & 12.10$\pm$0.16  \\
	LGP-NMT+	     & 29.06$\pm$0.25 &  77.57$\pm$0.24 & 12.09$\pm$0.27 \\ \hdashline
	SEQ	 			& 28.60$\pm$0.24 & 77.39$\pm$0.15 & 12.15$\pm$0.12 \\
	UNI	 				& 28.25$\pm$0.35  &  77.13$\pm$0.20 & 12.37$\pm$0.08 \\
	DEP 				& 26.83$\pm$0.38  &  76.05$\pm$0.22 & 13.33$\pm$0.23 \\ \hline
    \end{tabular}
}
    \caption{Evaluation on the development data using the medium training dataset (100,000 pairs).}
    \label{tb:comparison}
  \end{center}
\end{table}

Table~\ref{tb:comparison} shows the results of using the medium training dataset.
In contrast with using the small training dataset, LGP-NMT is slightly better than SEQ.
LGP-NMT significantly outperforms UNI, which shows that our adaptive learning is more effective than using the uniform graph weights.
By pre-training our model, LGP-NMT+ significantly outperforms SEQ in terms of the BLEU score.
Again, DEP performs the worst among all the models.

By using our beam search strategy, the Brevity Penalty (BP) values of our translation results are equal to or close to 1.0, which is important when evaluating the translation results using the BLEU scores.
A BP value ranges from 0.0 to 1.0, and larger values mean that the translated sentences have relevant lengths compared with the reference translations.
As a result, our BLEU evaluation results are affected only by the word $n$-gram precision scores.
BLEU scores are sensitive to the BP values, and thus our beam search strategy leads to more solid evaluation for NMT models.

\section{Results on Large Dataset}

\begin{table}[t]
  \begin{center}
{\small
    \begin{tabular}{l|ccc}
	\multicolumn{1}{r|}{B./R.}		& Single & +Averaging & +UnkRep \\ \hline
	LGP-NMT 		& 38.05/81.98 & 38.44/82.23 & 38.77/82.29 \\
	LGP-NMT+	& 38.75/82.13 & 39.01/82.40 & 39.37/82.48 \\ \hdashline
	SEQ	 		& 38.24/81.84 & 38.26/82.14 & 38.61/82.18 \\ \hline
    \end{tabular}
}
    \caption{BLEU (B.) and RIBES (R.) scores on the development data using the large training dataset.}
    \label{tb:dev_res}
  \end{center}

  \begin{center}
{\small
    \begin{tabular}{l|cc}
							& BLEU & RIBES \\ \hline
	LGP-NMT   	& 39.19  & 82.66  \\
	LGP-NMT+ 	& 39.42 & 82.83 \\ \hdashline
	SEQ 			& 38.96 & 82.18 \\ \hdashline
	Ensemble of the above three models & 41.18 & 83.40 \\ \hline
	\citet{kyotou}			& 38.20 & 82.39 \\
	\citet{neubig2015}		& 38.17 & 81.38 \\
	\citet{eriguchi2016wat} & 36.95 & 82.45 \\
	\citet{tree2str_2} & 36.58 & 79.65 \\
	\citet{zhu_wat} & 36.21 & 80.91 \\
	\citet{naver_wat} & 35.75 & 81.15 \\ \hline

    \end{tabular}
}
    \caption{BLEU and RIBES scores on the test data.}
    \label{tb:test_res}
  \end{center}
\end{table}

Table~\ref{tb:dev_res} shows the BLEU and RIBES scores on the development data achieved with the large training dataset.
Here we focus on our models and SEQ because UNI and DEP consistently perform worse than the other models as shown in Table~\ref{tb:comparison_small} and \ref{tb:comparison}.
The averaging technique and attention-based unknown word replacement~\cite{unkrep,hashimoto2016} improve the scores.
Again, we see that the translation scores of our model can be further improved by pre-training the model.

Table~\ref{tb:test_res} shows our results on the test data, and the previous best results summarized in \citet{wat16} and the WAT website\footnote{\url{http://lotus.kuee.kyoto-u.ac.jp/WAT/evaluation/list.php?t=1&o=1}.} are also shown.
Our proposed models, LGP-NMT and LGP-NMT+, outperform not only SEQ but also all of the previous best results.
Notice also that our implementation of the sequential model (SEQ) provides a very strong baseline, the performance of which is already comparable to the previous state of the art, even without using ensemble techniques.
The confidence interval $(p\leq0.05)$ of the RIBES score of LGP-NMT+ estimated by bootstrap resampling~\citep{bootstrap} is $(82.27, 83.37)$, and thus the RIBES score of LGP-NMT+ is significantly better than that of SEQ, which shows that our latent parser can be effectively pre-trained with the human-annotated treebank.

The sequential NMT model in \citet{kyotou} and the tree-to-sequence NMT model in \citet{eriguchi2016} rely on ensemble techniques while our results mentioned above are obtained using single models.
Moreover, our model is more compact\footnote{Our training time is within five days on a {\tt c4.8xlarge} machine of Amazon Web Service by our CPU-based C++ code, while it is reported that the training time is more than two weeks in \citet{kyotou} by their GPU code.} than the previous best NMT model in \citet{kyotou}.
By applying the ensemble technique to LGP-NMT, LGP-NMT+, and SEQ, the BLEU and RIBES scores are further improved, and both of the scores are significantly better than the previous best scores.

\subsection{Analysis on Translation Examples}

\begin{figure}[t]
	\begin{center}
    	\includegraphics[width=80mm]{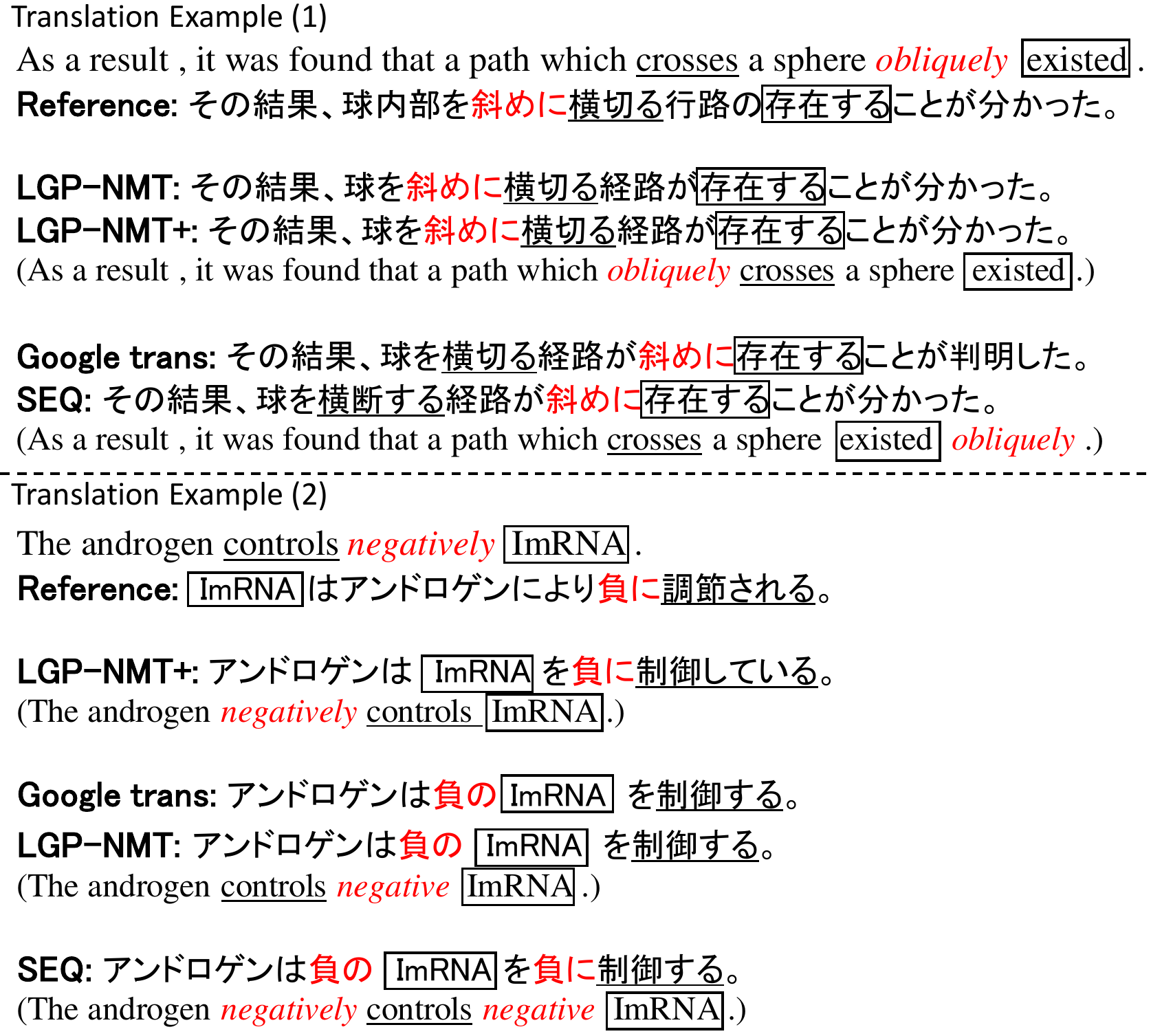}
    \end{center}
\vspace{-0.5cm}
\caption{English-to-Japanese translation examples for focusing on  the usage of adverbs.}
\label{fig:trans_1}
\end{figure}

Figure~\ref{fig:trans_1} shows two translation examples\footnote{These English sentences were created by manual simplification of sentences in the development data.} to see how the proposed model works and what is missing in the state-of-the-art sequential NMT model, SEQ.
Besides the reference translation, the outputs of our models with and without pre-training, SEQ, and Google Translation\footnote{The translations were obtained at \url{https://translate.google.com} in Feb. and Mar. 2017.} are shown.

\paragraph{Selectional Preference}
In the translation example (1) in Figure~\ref{fig:trans_1}, we see that the adverb ``obliquely'' is interpreted differently across the systems.
As in the reference translation, ``obliquely'' is a modifier of the verb ``crosses''.
Our models correctly capture the relationship between the two words, whereas Google Translation and SEQ treat ``obliquely'' as a modifier of the verb ``existed''.
This error is not a surprise since the verb ``existed'' is located closer to ``obliquely'' than the verb ``crosses''.
A possible reason for the correct interpretation by our models is that they can better capture long-distance dependencies and are less susceptible to surface word distances.
This is an indication of our models' ability of capturing domain-specific selectional preference that cannot be captured by purely sequential models.
It should be noted that simply using standard treebank-based parsers does not necessarily address this error, because our pre-trained dependency parser interprets that ``obliquely'' is a modifier of the verb ``existed''.

\paragraph{Adverb or Adjective}
The translation example (2) in Figure~\ref{fig:trans_1} shows another example where the adverb ``negatively'' is interpreted as an adverb or an adjective.
As in the reference translation, ``negatively'' is a modifier of the verb ``controls''.
Only LGP-NMT+ correctly captures the adverb-verb relationship, whereas ``negatively'' is interpreted as the adjective ``negative'' to modify the noun ``ImRNA'' in the translation results from Google Translation and LGP-NMT.
SEQ interprets ``negatively'' as both an adverb and an adjective, which leads to the repeated translations.
This error suggests that the state-of-the-art NMT models are strongly affected by the word order.
By contrast, the pre-training strategy effectively embeds the information about the POS tags and the dependency relations into our model.


\subsection{Analysis on Learned Latent Graphs}

\paragraph{Without Pre-Training}
We inspected the latent graphs learned by LGP-NMT.
Figure~\ref{fig:overview} shows an example of the learned latent graph obtained for a sentence taken from the development data of the translation task.
It has long-range dependencies and cycles as well as ordinary left-to-right dependencies.
We have observed that the punctuation mark ``.'' is often pointed to by other words with large weights.
This is primarily because the hidden state corresponding to the mark in each sentence has rich information about the sentence.

To measure the correlation between the latent graphs and human-defined dependencies, we parsed the sentences on the development data of the WSJ corpus and converted the graphs into dependency trees by Eisner's algorithm~\citep{eisner}.
For evaluation, we followed \citet{chen2014dep} and measured Unlabeled Attachment Score (UAS).
The UAS is 24.52\%, which shows that the implicitly-learned latent graphs are partially consistent with the human-defined syntactic structures.
Similar trends have been reported by \citet{deepmind} in the case of binary constituency parsing.
We checked the most dominant gold dependency labels which were assigned for the dependencies detected by LGP-NMT.
The labels whose ratio is more than 3\% are {\tt nn}, {\tt amod}, {\tt prep}, {\tt pobj}, {\tt dobj}, {\tt nsubj}, {\tt num}, {\tt det}, {\tt advmod}, and {\tt poss}.
We see that dependencies between words in distant positions, such as subject-verb-object relations, can be captured.

\begin{figure}[t]
	\begin{center}
    	\includegraphics[width=75mm]{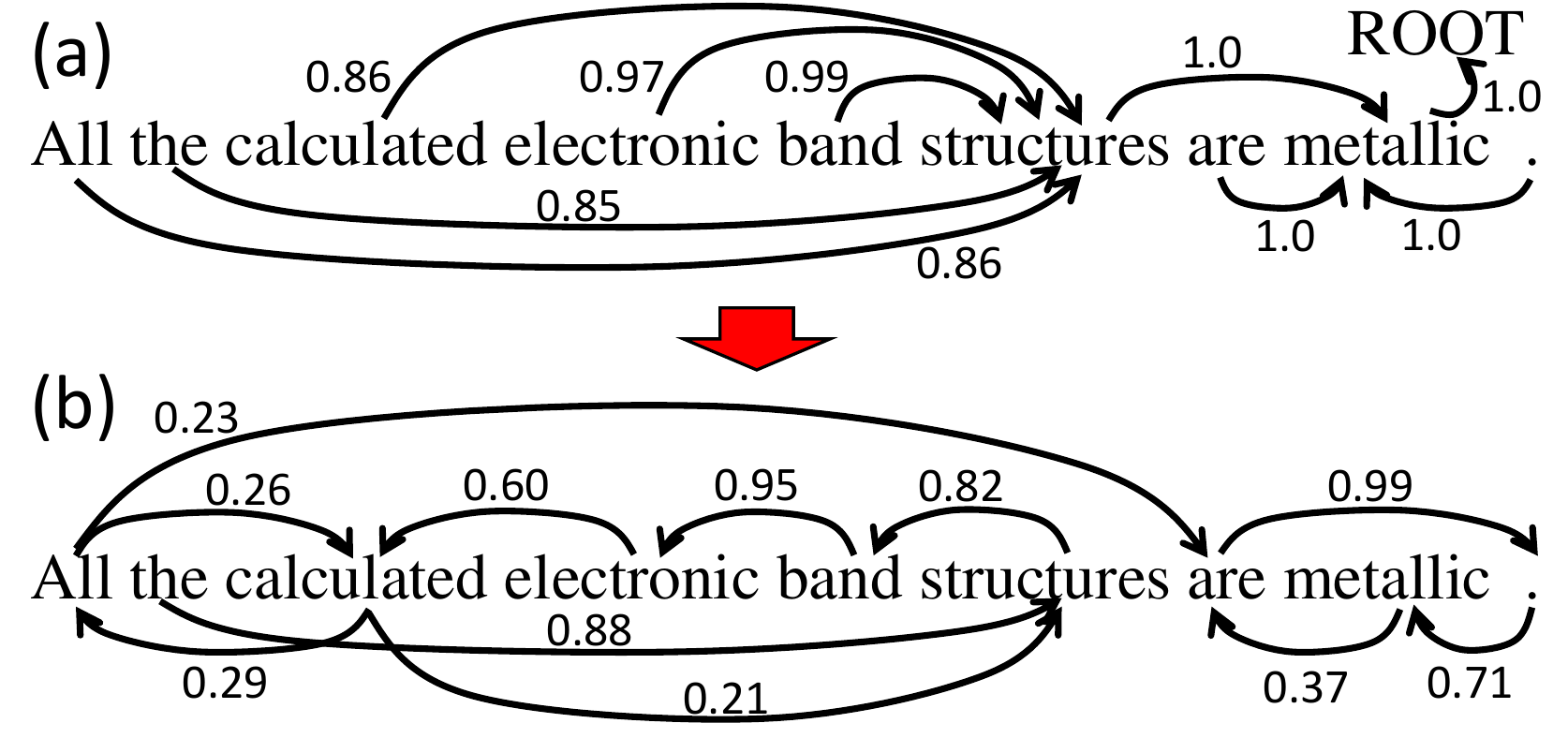}
    \end{center}
\vspace{-0.5cm}
\caption{An example of the pre-trained dependency structures (a) and its corresponding latent graph adapted by our model (b).}
\label{fig:pre}
\end{figure}

\paragraph{With Pre-Training}
We also inspected the pre-trained latent graphs.
Figure~\ref{fig:pre}-(a) shows the dependency structure output by the pre-trained latent parser for the same sentence in Figure~\ref{fig:overview}.
This is an ordinary dependency tree, and the head selection is almost deterministic; that is, for each word, the largest weight of the head selection is close to 1.0.
By contrast, the weight values are more evenly distributed in the case of LGP-NMT as shown in Figure~\ref{fig:overview}.
After the overall NMT model training, the latent parser is adapted to the translation task, and Figure~\ref{fig:pre}-(b) shows the adapted latent graph.
Again, we can see that the adapted weight values are also distributed and different from the original pre-trained weight values, which suggests that human-defined syntax is not always optimal for the target task.

The UAS of the pre-trained dependency trees is 92.52\%\footnote{The UAS is significantly lower than the reported score in \citet{jmt_emnlp}. The reason is described in Section~\ref{subsec:pre-training}.}, and that of the adapted latent graphs is 18.94\%.
Surprisingly, the resulting UAS (18.94\%) is lower than the UAS of our model without pre-training (24.52\%).
However, in terms of the translation accuracy, our model with pre-training is better than that without pre-training.
These results suggest that human-annotated treebanks can provide useful prior knowledge to guide the overall model training by pre-training, but the resulting sentence structures adapted to the target task do not need to highly correlate with the treebanks.


\section{Related Work}
While initial studies on NMT treat each sentence as a sequence of words~\citep{attention,luong2015,seq2seq}, researchers have recently started investigating into the use of syntactic structures in NMT models~\citep{graph_conv,chen2017tree2seq,eriguchi2016wat,eriguchi2016,eriguchi2017,li2017_syn,sennrich_wmt,guide,yang2017tree2seq}.
In particular, \citet{eriguchi2016} introduced a tree-to-sequence NMT model by building a tree-structured encoder on top of a standard sequential encoder, which motivated the use of the dependency composition vectors in our proposed model.
Prior to the advent of NMT, the syntactic structures had been successfully used in statistical machine translation systems~\citep{tree2str_2,tree2str_1}.
These syntax-based approaches are pipelined; a syntactic parser is first trained by supervised learning using a treebank such as the WSJ dataset, and then the parser is used to automatically extract syntactic information for machine translation.
They rely on the output from the parser, and therefore parsing errors are propagated through the whole systems.
By contrast, our model allows the parser to be adapted to the translation task, thereby providing a first step towards addressing ambiguous syntactic and semantic problems, such as domain-specific selectional preference and PP attachments, in a task-oriented fashion.

Our model learns latent graph structures in a source-side language.
\citet{eriguchi2017} have proposed a model which learns to parse and translate by using automatically-parsed data.
Thus, it is also an interesting direction to learn latent structures in a target-side language.

As for the learning of latent syntactic structure, there are several studies on learning task-oriented syntactic structures.
\citet{deepmind} used a reinforcement learning method on shift-reduce action sequences to learn task-oriented binary constituency trees.
They have shown that the learned trees do not necessarily highly correlate with the human-annotated treebanks, which is consistent with our experimental results.
\citet{socher2011} used a recursive autoencoder model to greedily construct a binary constituency tree for each sentence.
The autoencoder objective works as a regularization term for sentiment classification tasks.
Prior to these deep learning approaches, \citet{dekai} presented a method for {\it bilingual parsing}.
One of the characteristics of our model is directly using the soft connections of the graph edges with the real-valued weights, whereas all of the above-mentioned methods use one best structure for each sentence.
Our model is based on dependency structures, and it is a promising future direction to jointly learn dependency and constituency structures in a task-oriented fashion.

Finally, more related to our model, \citet{yoon2017kim} applied their {\it structured attention networks} to a Natural Language Inference (NLI) task for learning dependency-like structures.
They showed that pre-training their model by a parsing dataset did not improve accuracy on the NLI task.
By contrast, our experiments show that such a parsing dataset can be effectively used to improve translation accuracy by varying the size of the dataset and by avoiding strong overfitting.
Moreover, our translation examples show the concrete benefit of learning task-oriented latent graph structures.

\section{Conclusion and Future Work}
We have presented an end-to-end NMT model by jointly learning translation and source-side latent graph representations.
By pre-training our model using treebank annotations, our model significantly outperforms both a pipelined syntax-based model and a state-of-the-art sequential model.
On English-to-Japanese translation, our model outperforms the previous best models by a large margin.
In future work, we investigate the effectiveness of our approach in different types of target tasks.


\section*{Acknowledgments}
We thank the anonymous reviewers and Akiko Eriguchi for their helpful comments and suggestions.
We also thank Yuchen Qiao and Kenjiro Taura for their help in speeding up our training code.
This work was supported by CREST, JST, and JSPS KAKENHI Grant Number 17J09620.

\bibliography{bibtex.bib}
\bibliographystyle{emnlp_natbib}


\end{document}